\documentclass{article}

\usepackage{graphicx}
\usepackage{todonotes}
\usepackage{amsmath}

\usepackage{microtype}
\usepackage{graphicx}
\usepackage{subcaption}
\usepackage{booktabs}

\usepackage{color,soul}
\usepackage{caption}
\usepackage{pifont}
\usepackage{xcolor}
\usepackage{makecell}
\usepackage{bm}
\usepackage{natbib}

\usepackage[preprint]{neurips_2026}

\usepackage[utf8]{inputenc} % allow utf-8 input
\usepackage[T1]{fontenc}    % use 8-bit T1 fonts
\usepackage{hyperref}       % hyperlinks
\usepackage{url}            % simple URL typesetting
\usepackage{booktabs}       % professional-quality tables
\usepackage{amsfonts}       % blackboard math symbols
\usepackage{nicefrac}       % compact symbols for 1/2, etc.
\usepackage{microtype}      % microtypography
\usepackage{xcolor}         % colors

\title{Enhancing VLM Reward Models Through Structure-Aware Fine-Tuning}

\author{%
  Pyrros Koussios\\
  ETH Zürich\\
  \texttt{pkoussios@ethz.ch}\\
  \And
  Chenhao Li\\
  ETH Zürich\\
  \texttt{chhenhli@ethz.ch}
  \And
  Xin Chen\\
  ETH Zürich\\
  \texttt{chexin@ethz.ch}
  \And
  Andreas Krause\\
  ETH Zürich\\
  \texttt{krausea@ethz.ch}
}

\begin{document}

\maketitle

\begin{abstract}
Designing effective reward functions remains a major bottleneck in Reinforcement Learning (RL).
Recent work uses large foundation Vision-Language Models (VLMs) as reward models, computing text-observation similarity to bypass manual reward engineering.
Although promising, these rewards are often noisy and unreliable, limiting their direct utility during deployment.
We present Structure-Aware Fine-Tuning (SAFT), a simple, self-supervised method that refines these imperfect reward signals online without access to ground-truth supervision.
SAFT leverages intrinsic structural priors to regularize the VLM’s latent space via LoRA adapters.
We rigorously evaluate SAFT across a spectrum of base model capabilities to demonstrate its versatility.
Our results show that SAFT consistently denoises the reward landscape, yielding faster policy convergence and substantially improved alignment (EPIC distance) relative to the underlying base model, suggesting that failures can often be attributed to structural brittleness rather than semantic misunderstanding.
By replacing extensive human preference annotation with structural inductive biases inherent to the task, SAFT offers a scalable path for stabilizing text-conditioned RL and underscores the broader value of incorporating task structure as a general inductive bias.
\end{abstract}

\begin{figure}[h]
  \begin{center}
  \centerline{\includegraphics[width=0.96\columnwidth]{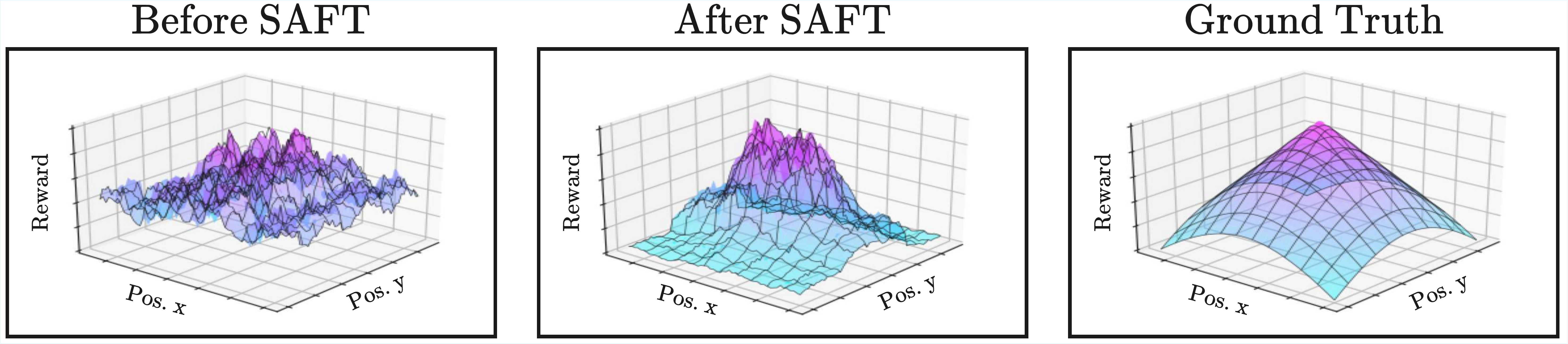}}
  \caption{Structure-Aware Fine-Tuning (SAFT) improves VLM reward models for RL by enforcing inherent structural priors, without ground-truth labels.}
  \label{fig:intro_fig}
  \end{center}
  \vskip -0.3in
\end{figure}

\section{Introduction}

\begin{figure*}[t]
  \begin{center}
  \centerline{\includegraphics[width=\linewidth]{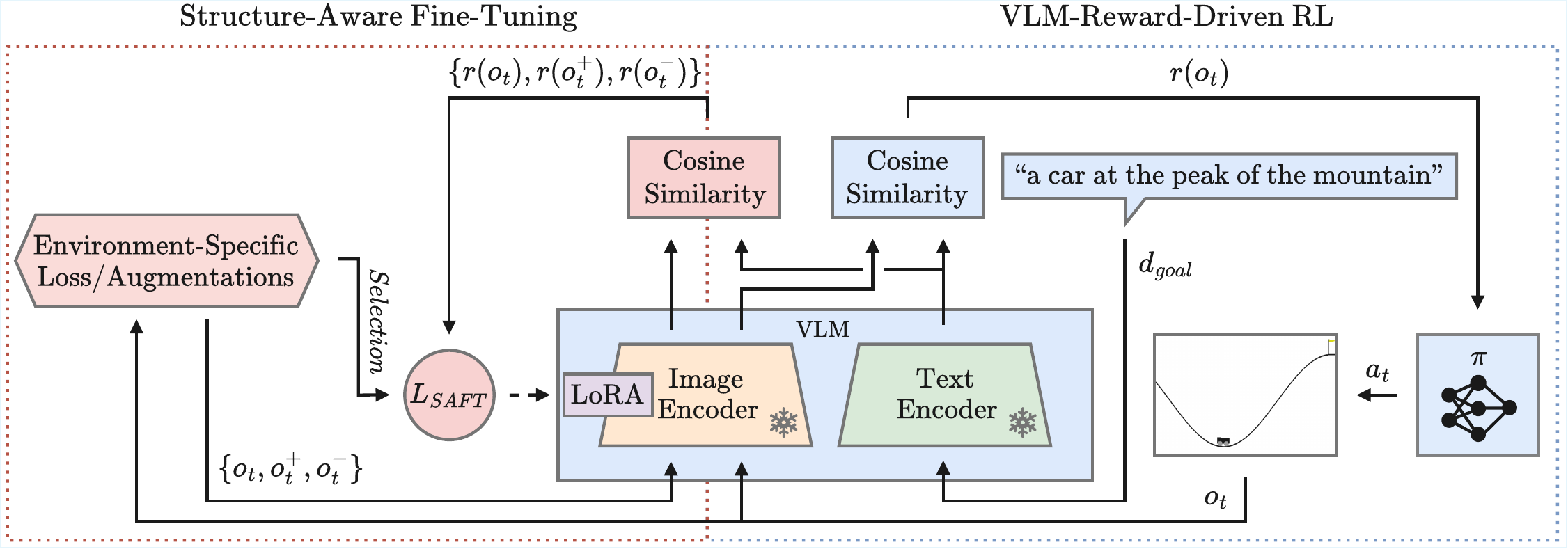}}
  \caption{Structure-aware fine-tuning adapts VLM reward models online without any ground-truth rewards by updating LoRA modules with environment-specific augmentations and auxiliary losses, leveraging structural priors to improve alignment, sample efficiency, and reduce human feedback.}
  \label{fig:saft}
  \vskip -0.3in
  \end{center}
\end{figure*}

Deep Reinforcement Learning (RL) has achieved strong results across many environments and applications \citep{levine2016endtoendtrainingdeepvisuomotor,starcraft}, but remains limited by the need for effective reward functions.
As tasks become more complex, designing state-based rewards often requires substantial human effort and iterative refinement \citep{rewardshaping}.
Reinforcement Learning from Human Feedback (RLHF) \citep{christiano2023deepreinforcementlearninghuman}, and more specifically Preference-based Reinforcement Learning (PbRL) \citep{Abdelkareem_2022}, offers an alternative by learning rewards from human preference comparisons.
However, due to the sparse nature of preference feedback, considerable human oversight is still required \citep{hejna2022fewshotpreferencelearninghumanintheloop}.

Recent work has explored leveraging Vision-Language Models (VLMs) as a bridge between human intent and reward specification \citep{wang2024rlvlmfreinforcementlearningvision, mahmoudieh2022zeroshot}.
VLMs with shared vision-language embedding spaces can act as drop-in reward models by computing similarity between textual goal embeddings and visual observation embeddings.
Prior work has used such rewards to derive dense signals from sparse rewards \citep{fu2024furlvisuallanguagemodelsfuzzy}, improve learning from demonstrations \citep{sontakke2023roboclipdemonstrationlearnrobot}, and train text-conditioned policies without explicit rewards or demonstrations \citep{rocamonde2024visionlanguage}.
However, a critical limitation of these approaches is their dependence on VLMs to generate consistent goal-distance estimates without fine-grained task understanding, demanding generalized high-fidelity scene perception capabilities that current models cannot achieve \citep{sontakke2023roboclipdemonstrationlearnrobot, rocamonde2024visionlanguage, fu2024furlvisuallanguagemodelsfuzzy}.

%To address these limitations, we present \emph{Structure-Aware Fine-Tuning (SAFT)}, demonstrating that VLM reward models can efficiently be fine-tuned in an online, self-supervised manner during reinforcement learning with minimal structural priors.
%Our approach enables solving otherwise intractable tasks, accelerates learning convergence, and improves alignment with ground-truth rewards when compared to non-finetuned baselines.
%Notably, SAFT delivers these benefits while dramatically reducing human labeling requirements relative to existing methods.

A central observation of this work is that such failures need not arise solely from semantic misunderstanding.
In many cases, VLM rewards contain useful task information but fail to respect simple structural properties of the environment.
Rather than correcting rewards with ground-truth labels or preference annotations, this suggests a broader alternative to human supervision in which reward models are improved by imposing structural priors already inherent to the task.
Building on this idea, we present \emph{Structure-Aware Fine-Tuning (SAFT)}, an online, self-supervised method for efficiently fine-tuning VLM reward models with task-inherent structural priors.
SAFT enables agents to solve previously intractable tasks, accelerates policy convergence, and improves alignment with ground-truth rewards relative to non-finetuned VLM baselines, while substantially reducing the amount of human labeling required compared to preference-based alternatives.

Our contributions are threefold. 
\textbf{(i)} First, we propose structure-aware fine-tuning using augmentations and auxiliary losses applied to VLM reward models during online training, achieving significant sample efficiency gains across a comprehensive set of environments.
\textbf{(ii)} Second, we demonstrate substantial reductions in human labeling requirements through preference-based learning evaluation, while allowing smaller models to solve previously intractable tasks.
\textbf{(iii)} Third, we show an improved alignment between fine-tuned VLM reward models and ground-truth reward functions, with benefits extending beyond online performance gains.

%Our experiments, performed across two standard control benchmarks and two high-dimensional robotic manipulation tasks, demonstrate that SAFT enhances VLM reward models by improving sample efficiency, reducing human labeling requirements, enabling smaller models to achieve performance comparable to larger variants, and overall bringing their predictions closer to the ground truth reward.

Our experiments across two standard control benchmarks and two high-dimensional robotic manipulation tasks show that incorporating task structure into VLM reward models is useful, regardless of the particular mechanism used to impose it.
SAFT instantiates this paradigm through CAL and RLR, improving sample efficiency, reducing human labeling requirements, enabling smaller models to approach the performance of larger variants, and bringing reward predictions closer to the ground truth.

\section{Background}
\subsection{Reinforcement Learning from human feedback}\label{ssec:rlhf}
Reinforcement Learning from Human Feedback (RLHF) \citep{christiano2023deepreinforcementlearninghuman} replaces a manually specified reward function with a learned reward model inferred from human feedback. In Preference-based Reinforcement Learning (PbRL) \citep{pbrl, christiano2023deepreinforcementlearninghuman}, this feedback takes the form of trajectory comparisons $(\tau_1, \tau_2, \tau_1 \succ \tau_2)$ used to train a parametric reward model $\hat{r}_\phi$.

PbRL commonly uses the Bradley-Terry likelihood
\begin{equation}\label{eq:pbrl_1}
\mathbb{P}_\phi[\tau_1 \succ \tau_2] =
\frac{\exp(\hat{R}_\phi(\tau_1))}
{\exp(\hat{R}_\phi(\tau_1)) + \exp(\hat{R}_\phi(\tau_2))},
\quad
\hat{R}_\phi(\tau) = \sum_{t=0}^{T-1} \hat{r}_\phi(s_t, a_t),
\end{equation}
and minimizes the negative log-likelihood
\begin{equation}\label{eq:pbrl_2}
L_{\text{pref}}(\phi) =
-\sum_{(\tau_1, \tau_2, \tau_1 \succ \tau_2)}
\log \mathbb{P}_\phi[\tau_1 \succ \tau_2].
\end{equation}
The learned reward is then used as a proxy objective for standard RL. In observation-based VLM reward modeling, $\hat{r}_\phi$ is instantiated over observations rather than explicit state-action pairs.

\subsection{CLIP-trained vision-language models}

Vision-Language Models (VLMs) learn joint representations of visual and textual modalities. CLIP-trained VLMs \citep{radford2021learningtransferablevisualmodels} use dual encoders: a vision encoder $f_{\text{image}}$ and a text encoder $f_{\text{text}}$. Given an image $I$ and text $T$, the encoders produce embeddings $v = f_{\text{image}}(I)$ and $w = f_{\text{text}}(T)$ in a shared $d$-dimensional space. Training maximizes cosine similarity for matching pairs and minimizes it for non-matching pairs via the symmetric cross-entropy loss:
\begin{equation}
L_{\text{CLIP}} = -\frac{1}{2N} \sum_{i=1}^{N} \left[
\log \frac{\exp(v_i \cdot w_i / \kappa)}{\sum_{j=1}^{N} \exp(v_i \cdot w_j / \kappa)}
+
\log \frac{\exp(v_i \cdot w_i / \kappa)}{\sum_{j=1}^{N} \exp(v_j \cdot w_i / \kappa)}
\right]
\end{equation}
where $\kappa$ is a temperature parameter and $N$ the batch size. Trained jointly on large-scale image-text pairs without task-specific supervision, the shared embedding space enables zero-shot transfer by comparing image embeddings with textual prompts.

\subsection{VLM reward models}
Consider a goal-based RL task specified by a textual goal description \(d_{\text{goal}}\). At each time step \(t\), the VLM encodes the observation \(o_t\) and goal text as \(\bm{v}_t = f_{\text{image}}(o_t)\) and \(\bm{w} = f_{\text{text}}(d_{\text{goal}})\). The reward is then given by their cosine similarity:
\begin{equation}
r_t = \cos(\bm{v}_t, \bm{w}) =
\frac{\bm{v}_t \cdot \bm{w}}{\|\bm{v}_t\| \|\bm{w}\|}.
\end{equation}

This reward can replace the true reward in standard RL, enabling the training of text-conditioned policies without modification to the underlying algorithm. Following prior work \citep{sontakke2023roboclipdemonstrationlearnrobot, rocamonde2024visionlanguage, fu2024furlvisuallanguagemodelsfuzzy}, policy learning still uses the underlying state \(s\), while the observation-based VLM reward acts as the task reward.

\section{Structure-aware fine-tuning for VLM reward models}
\subsection{Structure-aware fine-tuning}

Previous work has shown that reward models derived from foundational VLMs often fail catastrophically, with weak scene understanding and reward signals that collapse under even minor visual changes \citep{rocamonde2024visionlanguage, fu2024furlvisuallanguagemodelsfuzzy}.
These brittleness issues make fine-tuning essential for building a stable and accurate reward landscape.
Injecting inductive bias is a natural way to encourage consistency, motivating our central question of whether task-inherent structural priors can mitigate VLM reward failures without ground-truth labels during fine-tuning.

Accordingly, we introduce Structure-Aware Fine-Tuning (SAFT), a simple LoRA-based \citep{hu2021loralowrankadaptationlarge} procedure that strengthens VLM reward models using auxiliary self-supervised objectives that encode the invariances and relational patterns of the target environment.
SAFT implements this idea through two complementary auxiliary objectives: one that enforces invariance to task-preserving transformations and another that promotes proportionality between state changes and reward changes.
Together, these losses yield reward models that respect environment symmetries, are less noisy, and align more closely with the true reward.
We freeze the base VLM and insert LoRA adapters into the image encoder, fine-tuning them online with one of the two objectives described below. Figure~\ref{fig:saft} shows this process. Hardware requirements and wall-clock time increases due to fine-tuning are detailed in Appendix~\ref{app:hardware}.

\subsubsection{Invariance via Contrastive Augmentation Loss (CAL)}\label{sssec:CAL}

\begin{figure}[t]
    \centering
    \includegraphics[width=1.0\textwidth]{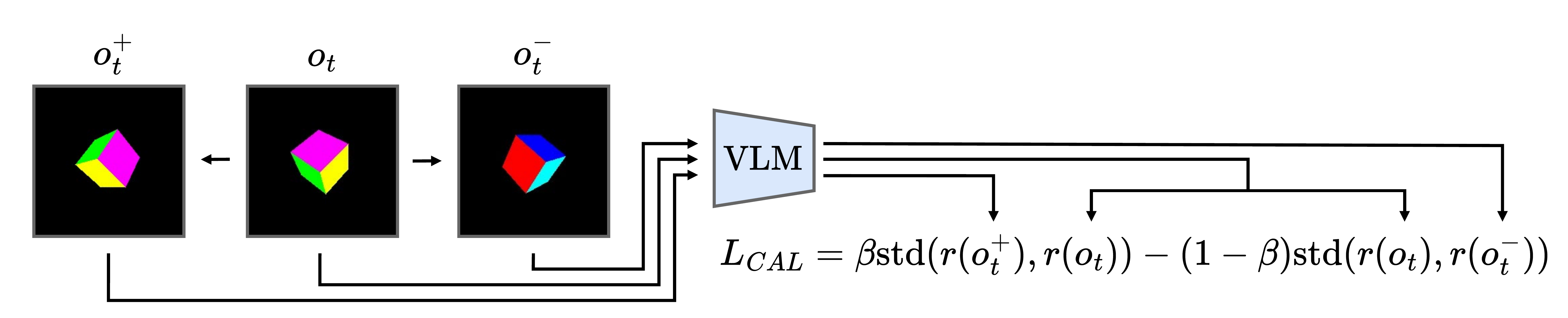}
    \caption{Contrastive Augmentation Loss, defined in \eqref{eq:CAL}, rewards invariance for positive augmentations while increasing separation from negative examples, preserving task semantics and avoiding representational collapse. 
  Visualization pertains to the ReposeCube environment.}
    \captionsetup{justification=centering}
    \label{fig:cal}
    \vskip -0.1in
\end{figure}

Reinforcement learning environments often exhibit structural symmetries that should be preserved in the reward function.
For instance, in CartPole, a pole tilted 15 degrees left or right from vertical corresponds to equivalent states that should yield similar rewards.

We exploit these symmetries through task-specific augmentations.
Positive augmentations $\{o_{t,i}^+\}_{i=1}^p$ are a set of $p$ transformations that should produce similar or equal rewards (e.g., horizontal flips in CartPole).
Contrastive examples $\{o_{t,j}^-\}_{j=1}^n$ are a set of $n$ semantically distinct states that should yield different rewards.
These can include hard negatives (e.g., vertical flips in CartPole) or softer negatives such as other time steps within the trajectory.

Given observation $o_t$, our \emph{Contrastive Augmentation Loss (CAL)} enforces reward invariance to positive transformations while maintaining discriminative capacity:
\begin{equation}\label{eq:CAL}
L_{\text{CAL}}
= \beta\,\mathrm{std}\!\big(r(o_t), r(o_{t,1}^+), \ldots, r(o_{t,p}^+)\big)
- (1-\beta)\,\mathrm{std}\!\big(r(o_t), r(o_{t,1}^-), \ldots, r(o_{t,n}^-)\big)
\end{equation}
where $r(o) = \cos(f_{\text{image}}(o), w)$, and \(\beta \in [0,1]\) is a tunable weighting term. The first term minimizes variance among positive augmentations, enforcing invariance to semantically equivalent transformations.
The second term maximizes variance among contrastive examples, ensuring the model maintains the ability to distinguish states and prevents representational collapse where all states receive similar rewards.
For all experiments, we utilize $\beta=0.5$, providing equal weighting to both terms.
Ablation studies regarding this value are detailed in Appendix~\ref{app:cal_weighting}.

We find that most environments naturally contain states with identical reward values, for example when camera viewpoints change, when background noise is present, or when states are equally related to the goal along different dimensions, making our augmentation-based approach broadly applicable across diverse tasks. 

\subsubsection{Proportionality via Reward Lipschitz Regularization (RLR)}

\begin{figure}[t]
  \begin{center}
  \centerline{\includegraphics[width=\columnwidth]{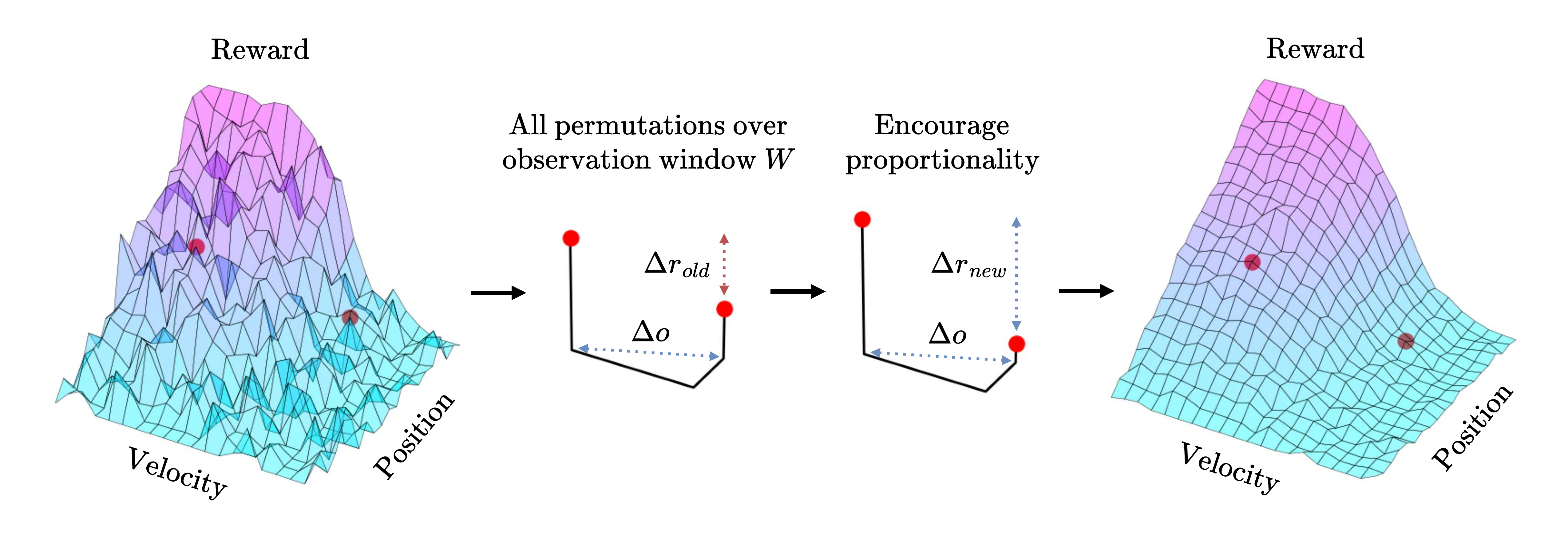}}
  \caption{Reward Lipschitz Regularization (RLR), defined in \eqref{eq:RLR}, encourages proportional changes between observations and rewards, punishing large spikes and inconsistencies within the reward model.
  Visualization pertains to the MountainCar environment.}
  \label{fig:rlr}
  \end{center}
  \vskip -0.3in
\end{figure}

%In goal-directed environments, changes in reward should often be proportional to changes in observations, reflecting the intuition that states closer to the goal should receive higher rewards.
In goal-directed environments, rewards often change smoothly with states within a local region, reflecting the intuition that states closer to the goal tend to have higher rewards.
We encourage this structural prior through \emph{Reward Lipschitz Regularization (RLR)}, using a soft loss-based approach rather than the hard constraints typical of traditional Lipschitz methods \citep{gouk2020regularisationneuralnetworksenforcing}.
A visualization of the effects of this auxiliary loss is shown in Figure~\ref{fig:rlr}.

Given a window of $W$ consecutive states $\{s_i\}_{i=1}^W$, we compute the scalar mean reward $\bar{r}$ and the element-wise mean state $\bar{s}$. We then define normalized rewards as $\tilde{r}_i = r(o_i) / \bar{r}$ and normalized states as $\tilde{s}_i = s_i / \bar{s}$, where the division is also applied element-wise. With this, we define the Lipschitz regularization loss as:
\begin{equation}\label{eq:RLR}
L_{\text{RLR}} = \sum_{i=1}^W \sum_{j \neq i} \left( \frac{\|\tilde{r}_i - \tilde{r}_j\|_2}{\|\tilde{s}_i - \tilde{s}_j\|_2 + \epsilon} - 1 \right)^2
\end{equation}
where $\epsilon > 0$ avoids division by zero.
This loss penalizes deviations from proportionality between state distances and reward differences, encouraging the reward function to respect the underlying geometric structure of the state space.
The window size $W$ determines the temporal scope over which proportionality is enforced.
Larger windows impose stronger structural priors by requiring consistency across longer temporal horizons.
Ablation studies regarding this value are detailed in Appendix~\ref{app:rlr_window_size}.

Although accurately measuring distances between states remains an open research problem, we adopt the L2 distance because of its simplicity and broad applicability within state-based RL.
This regularization is effective across diverse environments, particularly where proximity in the state space reflects progress toward goals.
In settings where the underlying state is inaccessible, or in high-dimensional environments where noise dominates the state and the assumptions of RLR are difficult to satisfy, CAL is generally more effective.

\section{Experiments}

We conduct a variety of experiments to evaluate the efficacy of SAFT.
%Our evaluation examines which structural priors best match different environment types, whether online fine-tuning improves sample efficiency during policy learning, how much human labeling effort can be reduced when viewed through a preference-based learning lens, and to what extent our fine-tuned reward models align with ground-truth rewards.
Our evaluation tests whether imposing appropriate task structure on VLM reward models improves reward quality, policy learning, and alignment. We examine which structural priors match different environment types, whether online fine-tuning improves sample efficiency, how much human labeling effort is reduced relative to preference-based learning, and whether the resulting rewards align more closely with ground truth.

\subsection{Matching structural priors to environments}

We evaluate our method across four goal-based RL environments: CartPole and MountainCar from the classic control suite \citep{brockman2016openaigym}, where VLM reward models have shown prior success, and modified versions of Reach (Franka) and ReposeCube (Allegro) from Isaac Lab \citep{mittal2023orbit}, representing realistic and high-dimensional robotic manipulation tasks.
We highlight that the ReposeCube task, with its 72-dimensional observation space and 16-dimensional action space, represents a significant challenge on par with complex humanoid locomotion.
All learning is conducted solely with the rewards provided by the VLM.
We additionally note that we purposely omit the Humanoid task used in prior work \citep{rocamonde2024visionlanguage} as it lacks a well-defined ground-truth reward function.
This mandates a reliance on subjective human-labeled success, which precludes a rigorous quantitative evaluation of policy performance.

The choice of auxiliary loss is determined by the structural properties available in each environment, as summarized in Table~\ref{tab:env_matching}. We use RLR when local state distances provide meaningful information about reward differences, and CAL when the environment admits task-preserving transformations that should leave the reward unchanged. Under this criterion, CartPole supports both CAL and RLR, MountainCar supports only RLR, and Reach and ReposeCube support CAL through rotational invariances. Exact details behind these matchings are provided in Appendix~\ref{app:env_specific_config}. These matchings are not meant to exhaust the space of possible priors, but to test whether simple task-inherent structure can serve as useful supervision for VLM reward adaptation.

\newcommand{\cmark}{\textcolor[rgb]{0,0.39,0}{\ding{51}}}%
\newcommand{\xmark}{\textcolor[rgb]{0.54,0,0}{\ding{55}}}%

\begin{figure}[t]
\centering
\begin{minipage}[c]{0.26\linewidth}
  \centering
  \includegraphics[width=\linewidth]{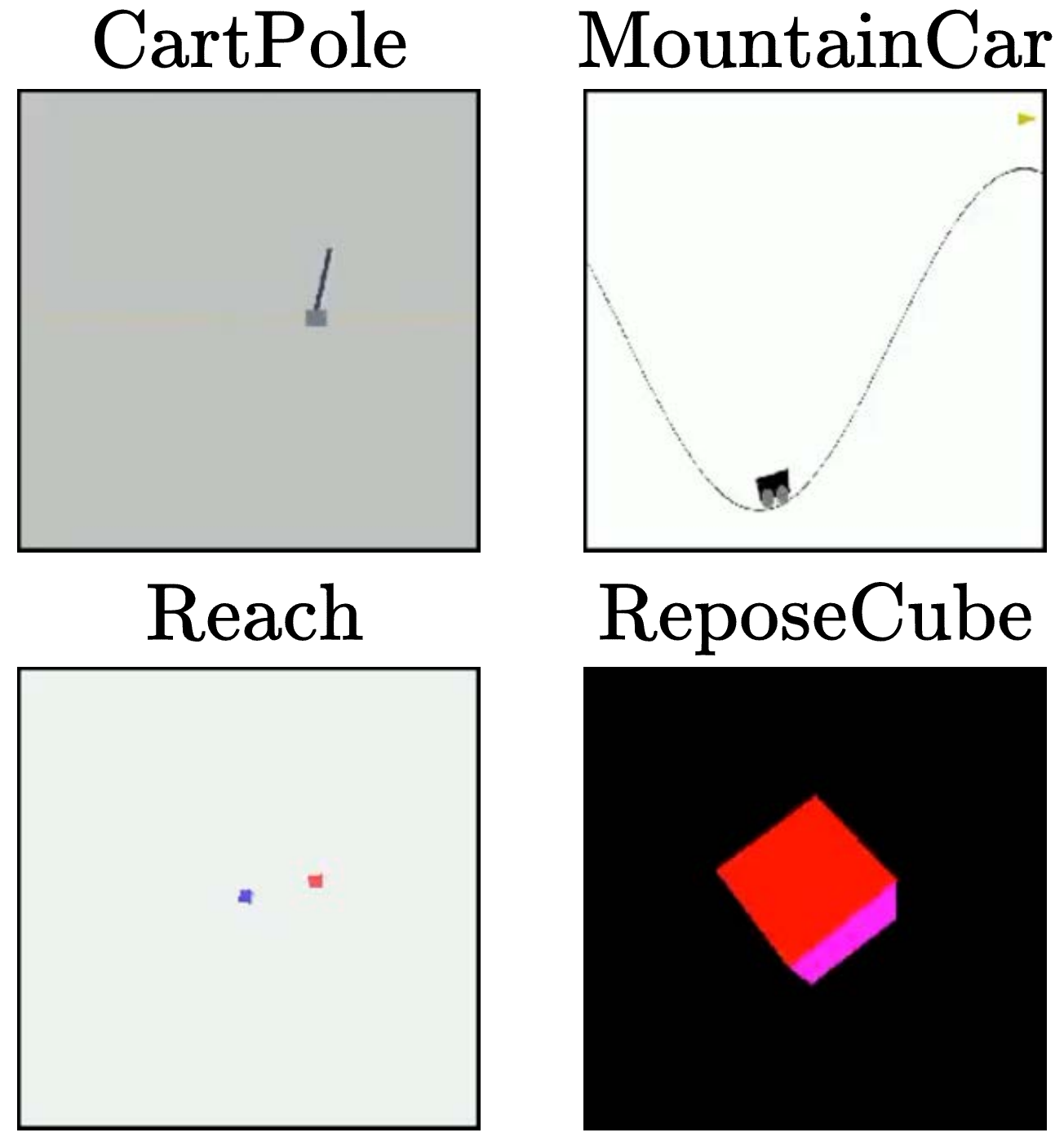}
  \captionof{figure}{Visualizations of the four environments.}
\end{minipage}
\hfill
\begin{minipage}[c]{0.7\linewidth}
    \centering
    \begin{tabular}{c c c c c}
    \toprule
    & \multicolumn{2}{c}{\textit{Classic Control}} & \multicolumn{2}{c}{\textit{Isaac Lab}} \\
    \cmidrule(lr){2-3} \cmidrule(lr){4-5} & CartPole & MountainCar & Reach & ReposeCube \\ \addlinespace[0.5em]
    CAL & \cmark & \xmark & \cmark & \cmark \\
    \addlinespace[0.5em]
    RLR & \cmark & \cmark & \xmark & \xmark \\
    \bottomrule
    \end{tabular}
    \captionsetup{skip=15pt}
    \captionof{table}{Applicability of Contrastive Augmentation Loss (CAL) and Reward Lipschitz Regularization (RLR) across all four environments.}
    \label{tab:env_matching}
\end{minipage}
\vskip -0.3in
\end{figure}

\subsection{Policy performance}\label{sec:policy_performance}

We first evaluate the sample efficiency of SAFT compared to baseline methods without fine-tuning, as well as goal-baseline regularization \citep{rocamonde2024visionlanguage} from prior work.
We omit direct comparisons to RoboCLIP \citep{sontakke2023roboclipdemonstrationlearnrobot} and FuRL \citep{fu2024furlvisuallanguagemodelsfuzzy} as they operate under fundamentally different assumptions: RoboCLIP requires video-based temporal information for trajectory-level rewards, while FuRL relies on privileged ground-truth sparse reward signals for fine-tuning.
Consequently, \citet{rocamonde2024visionlanguage} remains the only methodologically consistent baseline for our unsupervised, image-based setting.

We emphasize that our primary evaluation metric is the \emph{relative} improvement yielded by SAFT over a base model, rather than absolute task performance.
As SAFT is designed to refine existing reward signals, its utility naturally depends on the quality of the underlying VLM, a relationship we analyze across a broader spectrum in Section~\ref{sec:human_labeling_efficiency}.

However, to capture the nuances of online training, such as convergence speed, stability, and sample efficiency, that scalar summary metrics cannot convey, we focus our main policy analysis on a representative regime where the base VLM is functional but imperfect.
We initialize the VLM at a performance threshold where it can partially solve the task, allowing us to explicitly demonstrate how SAFT accelerates learning and stabilizes the reward landscape compared to the base model.
This initialization is performed through pretraining using the ground-truth reward as detailed in Appendix~\ref{app:pretraining}.
This controlled setting isolates the benefits of our proposed method while also allowing us to use a computationally efficient $86$ million parameter ViT-B-16 model \citep{ilharco_gabriel_2021_5143773} that achieves performance comparable to larger models without the associated computational overhead.
Critically, this initialization serves only to establish a consistent starting point for evaluation.
We measure the \emph{marginal utility} of our structural priors relative to the equally pretrained base model, with no ground-truth information available to the SAFT agent during the fine-tuning phase.

Our fine-tuning procedure uses LoRA adapters with $180$K parameters, representing $0.2$\% of the total model parameters, consistent with standard practice. 
We evaluate performance across five seeds.

\begin{figure*}
    \begin{center}
    \centerline{\includegraphics[width=0.99\linewidth]{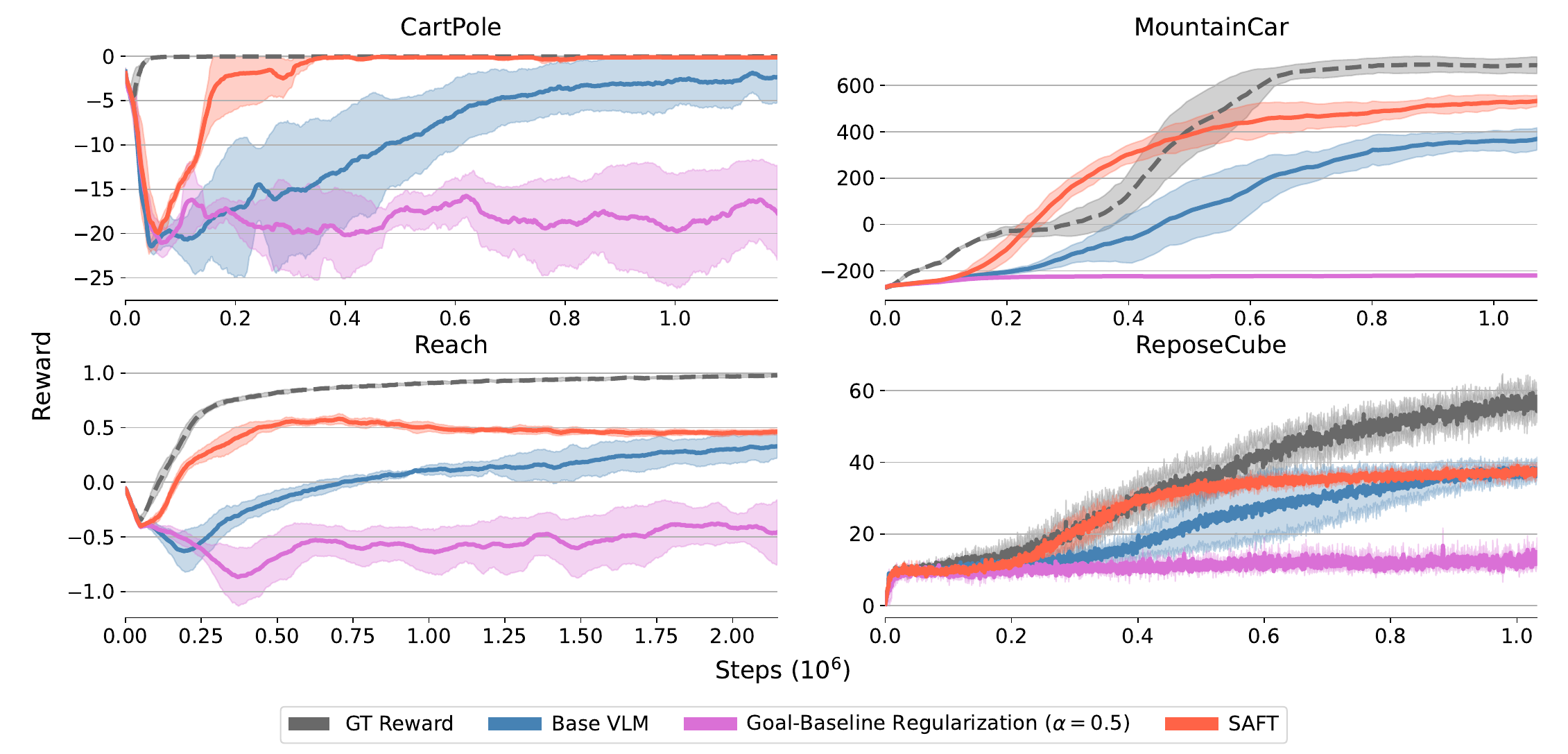}}
    \caption{Policy learning curves comparing SAFT against the base VLM, goal-baseline regularization \citep{rocamonde2024visionlanguage}, and Ground-Truth (GT) rewards across four environments. 
    SAFT demonstrates superior sample efficiency and convergence speed, substantially reducing the performance gap to GT rewards.
    Shaded regions indicate standard deviation across five seeds.}
    \label{fig:policy_performance}
    \end{center}
    \vskip -0.2in
\end{figure*}

Figure~\ref{fig:policy_performance} shows that SAFT consistently outperforms the baselines across environments by correcting reward inconsistencies and producing smoother reward landscapes.
When the base VLM fails to reach ground-truth performance, SAFT often achieves a higher final reward by improving shaping in high-reward regions, helping the policy identify task-success states.
Overall, SAFT substantially narrows the gap between VLM-derived and ground-truth rewards.

\subsection{Reward model alignment}\label{sec:offline_eval}

\begin{figure}[t]
\centering
\begin{minipage}[c]{0.60\linewidth}
  \centering
  \begin{tabular}{l c c}
    \toprule
    Environment & \multicolumn{2}{c}{EPIC Distance $\downarrow$} \\
    \cmidrule(lr){2-3}
                 & Before SAFT & After SAFT \\
    \midrule
    \textit{Classic Control} \\
    \quad CartPole    & $0.4306$ & $\mathbf{0.2782}\,\pm 0.0205$ \\
    \quad MountainCar & $0.3106$ & $\mathbf{0.2330}\,\pm\,0.0074$ \\
    \addlinespace[0.5em]
    \textit{Isaac Lab} \\
    \quad Reach       & $0.3443$ & $\mathbf{0.3129}\,\pm\,0.0069$ \\
    \quad ReposeCube  & $0.2963$ &$\mathbf{0.2605}\,\pm 0.0098$ \\
    \bottomrule
  \end{tabular}
  \captionsetup{skip=10pt}
  \captionof{table}{EPIC distance between VLM reward models and the ground-truth reward. SAFT reduces EPIC distance, indicating closer alignment to the ground truth. Post-fine-tuning results report mean and one standard deviation across five seeds.}
  \label{tab:epic_distance}
\end{minipage}
\hfill
\begin{minipage}[c]{0.36\linewidth}
  \centering
  \includegraphics[width=\linewidth]{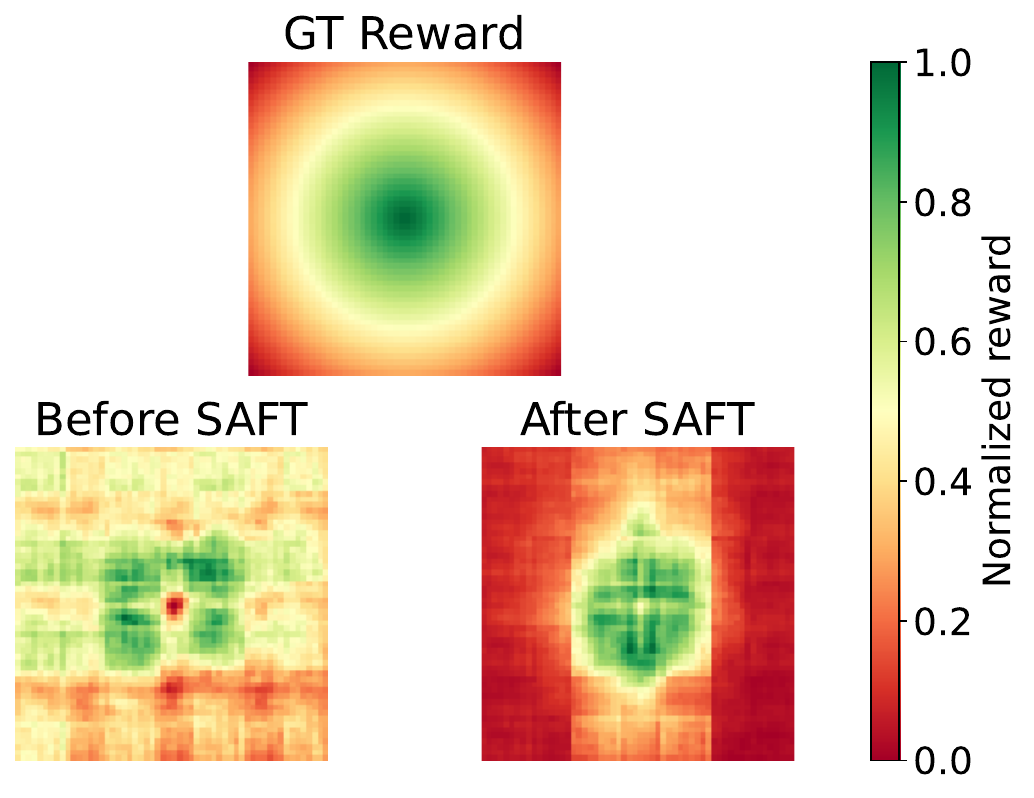}
  \captionof{figure}{Reach reward landscapes predicted by the VLM with and without SAFT, showing improved alignment to the ground truth under SAFT.}
  \label{fig:reach_rewards}
\end{minipage}
\vskip -0.25in
\end{figure}

Next, we conduct offline evaluation to assess how well our fine-tuned VLMs align with ground-truth rewards compared to their unmodified counterparts.
We employ the Equivalent-Policy Invariant Comparison (EPIC) \citep{gleave2021quantifyingdifferencesrewardfunctions} distance as our evaluation metric, which measures differences between reward functions by comparing their canonical forms while removing the effects of reward shaping and scaling.
This metric captures only the differences that affect optimal behavior, making it ideally suited for our analysis.
For this analysis, we employ the base and fine-tuned VLM checkpoints from Section~\ref{sec:policy_performance}.

Table~\ref{tab:epic_distance} presents the EPIC distance measurements on the different environments.
Across all environments, fine-tuned models demonstrate substantially lower EPIC distances compared to their non-fine-tuned counterparts, indicating closer alignment with ground truth reward functions.
These alignment improvements suggest that structural fine-tuning improves the reward model itself, rather than merely producing policies that exploit incidental shaping effects.
Figure~\ref{fig:reach_rewards} shows VLM-predicted rewards before and after applying SAFT. Further results using additional VLM backbones are provided in Appendix~\ref{app:backbone}.

\subsection{Human labeling efficiency}\label{sec:human_labeling_efficiency}

Finally, we evaluate SAFT against preference-based reinforcement learning (PbRL), one of the most commonly used approaches for learning reward models when manually specified rewards are unavailable or unreliable \citep{christiano2023deepreinforcementlearninghuman}.
We implement a PbRL baseline under the same setup as SAFT, freezing the base VLM and updating only LoRA adapters in the image encoder.
After each rollout, binary preference queries are generated and the reward model is updated online using the Bradley-Terry objective from Section~\ref{ssec:rlhf}.
We then measure how many preference queries are required for PbRL to match the performance achieved by SAFT.
Implementation details are provided in Appendix~\ref{app:pbrl_details}.

Table~\ref{tab:comparisons_saved} reports the number of preference queries required by PbRL to match SAFT performance.
Across environments, SAFT replaces several hundred to several thousands of binary comparisons with a one-time specification of structural priors.
This both reduces human labeling effort and avoids the accumulated noise and inconsistency of online human feedback.

\begin{figure*}
    \centering
    \includegraphics[width=0.95\textwidth]{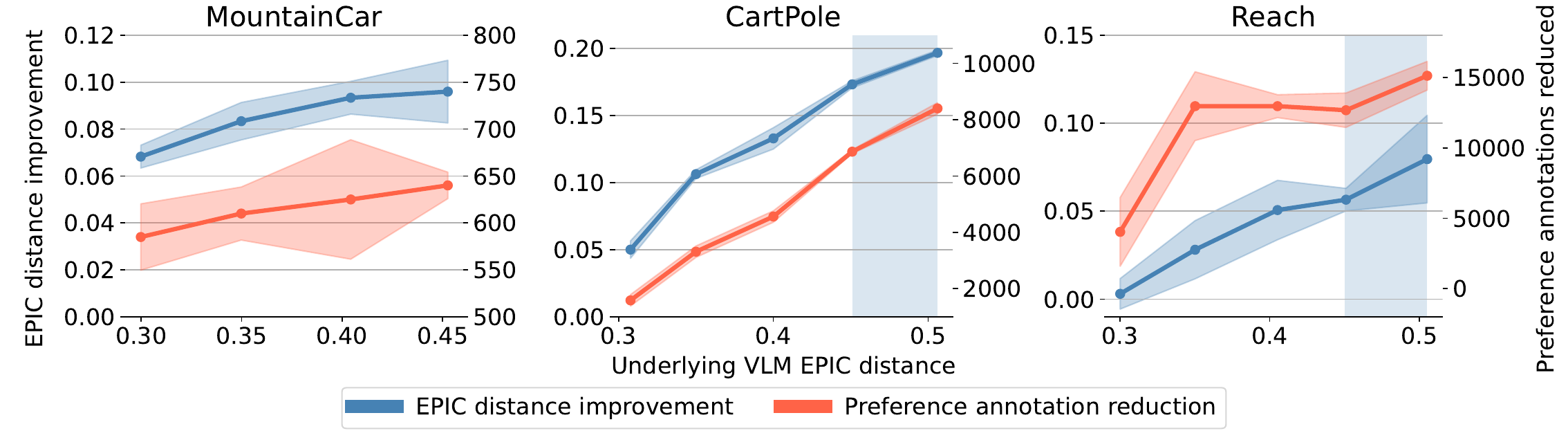}
    \caption{Utility of SAFT in relation to base VLM reward quality, averaged over two seeds.
    As the underlying VLM degrades, SAFT becomes increasingly useful. The critical regime, highlighted in blue, is where the base VLM fails yet SAFT enables successful training.}
    \label{fig:epic_vs_comparisons}
    \vskip -0.0in
\end{figure*}

\begin{table}[t]
\centering
\begin{tabular}{c c c c c}
\toprule
& \multicolumn{2}{c}{\textit{Classic Control}} & \multicolumn{2}{c}{\textit{Isaac Lab}} \\
\cmidrule(lr){2-3} \cmidrule(lr){4-5} & CartPole & MountainCar & Reach & ReposeCube \\ \addlinespace[0.5em]
Binary Comparisons Saved $\uparrow$ & $5\,608 \pm 309$ & $670 \pm 27$ & $13\,516 \pm 1\,682$ & $8\,755 \pm 1\,711$ \\
\bottomrule
\end{tabular}
\captionsetup{skip=10pt}
\captionof{table}{Number of binary preference comparisons eliminated by using SAFT instead of human feedback, rounded to the closest integer. SAFT substantially reduces required human effort.
Values show mean and one standard deviation across five seeds.}
\label{tab:comparisons_saved}
\vskip -0.3in
\end{table}

As stated in Section~\ref{sec:policy_performance}, the effectiveness of our method depends on the initial quality of the underlying VLM.
An oracle VLM that perfectly models the ground-truth reward would render our improvements less impactful.
To characterize this relationship, we evaluate binary feedback savings as a function of underlying VLM strength by applying SAFT to VLMs with varying levels of pretraining, as presented in Figure~\ref{fig:epic_vs_comparisons}. 

As the VLM reward model deviates from the ground truth, SAFT becomes increasingly valuable by correcting inconsistencies and reducing noise in the reward signal.
We observe that SAFT consistently improves performance across varying VLM strengths, lowering EPIC distance and effectively substituting for costly online human labeling.
Notably, for most environments, there exists a critical regime in which the base VLM alone cannot solve the task due to noisy scene understanding, yet SAFT enables successful learning by mitigating this noise.
Smaller variants within the same model family can fall into this regime and fail where larger counterparts succeed.
Our method bridges this gap (Appendix~\ref{app:smaller_models}).
Past a certain point, when the VLM is fundamentally misaligned with the task, our method cannot enable policy convergence, as the auxiliary losses operate without ground-truth supervision and are intended to improve consistency and reduce noise in the reward signal, but cannot compensate for a reward model that lacks any ability to interpret the scene.

\section{Related work}

\paragraph{Self-supervised auxiliary losses for reinforcement learning.} 

Self-supervised auxiliary losses have long improved generalization and policy performance in RL \citep{jaderberg2016reinforcementlearningunsupervisedauxiliary, shelhamer2017lossrewardselfsupervisionreinforcement, yarats2020improvingsampleefficiencymodelfree}.  
Image augmentations are especially effective for vision-based policies \citep{laskin2020reinforcementlearningaugmenteddata, yarats2021masteringvisualcontinuouscontrol}, and contrastive learning methods such as CURL have also proven effective \citep{srinivas2020curlcontrastiveunsupervisedrepresentations}. 

Another line of work has explored constraining networks to be Lipschitz continuous \citep{asadi2018lipschitzcontinuitymodelbasedreinforcement,
scaman2019lipschitzregularitydeepneural,
gouk2020regularisationneuralnetworksenforcing}.
We take a similar approach but regularize the reward function rather than enforcing hard constraints and use L2 distances between states as a baseline despite the existence of more sophisticated metrics \citep{jiang2020reinforcementlearninggoaldistancegradient,
myers2025learningtemporaldistancescontrastive}.

So far, these auxiliary losses have been applied to policy networks. To the best of our knowledge, we are the first to adapt them for online fine-tuning of VLM reward models, combining Lipschitz regularization and visual augmentations to improve reward quality.

\paragraph{Large pretrained models as reinforcement learning reward functions.} 

RLHF \citep{christiano2023deepreinforcementlearninghuman} aligns policies with human intent by training reward models from preference data, but collecting such data is costly.
Recent work instead uses foundation models for automated reward specification.
Early methods had LLMs generate reward signals and refine them with policy feedback \citep{kwon2023reward, song2023selfrefinedlargelanguagemodel}, while newer ones generate reward code \citep{xie2024textreward, ma2024eurekahumanlevelrewarddesign, li2025r}.
However, the unimodality of LLMs limits these approaches to relying on the indirectness of intermediary textual representations.

Multimodal VLMs offer a promising alternative.
Initial studies used them as success detectors \citep{cui2022foundationmodelsperformzeroshot, du2023visionlanguagemodelssuccessdetectors} or for generating preference labels in PbRL \citep{wang2024rlvlmfreinforcementlearningvision}, though these approaches reduce information to binary signals or require auxiliary models.
Others trained policies using embedding similarity \citep{mahmoudieh2022zeroshot}, but required large offline datasets.

Most closely related, \citet{rocamonde2024visionlanguage} and \citet{sontakke2023roboclipdemonstrationlearnrobot} use VLMs directly as reward models, with the former introducing goal-baseline regularization and the latter applying video-based models for motion goals.
\citet{fu2024furlvisuallanguagemodelsfuzzy} further show that fine-tuning with sparse ground-truth rewards improves dense reward quality.
Our work builds on these directions, but argues that task structure can serve as a direct source of supervision for improving VLM reward models, without requiring ground-truth rewards or preference labels during online fine-tuning.

\section{Limitations and future work}
Our primary limitation is that the auxiliary loss, the augmentations for CAL, and the window size for RLR are chosen on an environment by environment basis, which reduces generality.
However, this situation is not unique to our work.
Early visual RL began with seemingly ad hoc augmentation techniques \citep{sadeghi2017cad2rlrealsingleimageflight, Lee2020Network, cobbe2019quantifyinggeneralizationreinforcementlearning}, and over time, work such as RAD \citep{laskin2020reinforcementlearningaugmenteddata} revealed randomized crops as a strategy that generalized well.

In the same spirit, we introduce ground-truth-free auxiliary losses for fine-tuning VLM reward models, playing a role for reward learning analogous to early augmentation studies for policy learning.
SAFT shows that simple structural priors improve reward alignment, increase sample efficiency, and reduce human supervision.
We do not claim a general purpose solution, but we view SAFT as a step toward more automated and broadly applicable approaches.

A natural next step is to replace manual choices with methods that automatically discover and combine structural priors within a unified framework that subsumes our current objectives.
The text encoder remains frozen and underused, so future work could fine-tune it with contrastive goal descriptions and their textual negations to sharpen task understanding.
As tasks grow in complexity, hierarchical scene decomposition and compositional augmentation may be required, for example in manipulation tasks that demand both rotation invariant grasping and position sensitive placement.

\section{Conclusion}

In this work, we introduce Structure-Aware Fine-Tuning (SAFT), a simple method for adapting frozen vision-language models whose purpose it is to provide rewards for reinforcement learning.
SAFT incorporates inductive bias through simple structural priors, applied via LoRA adapters to reshape and denoise the reward signal.
This yields smoother and more consistent reward landscapes, faster policy convergence, closer alignment with the ground truth task reward, and the ability for smaller VLMs to solve tasks that otherwise require larger models.
%Our experiments across both classic control and robotic manipulation environments show that the brittleness and misalignment issues of general foundation models can be effectively mitigated by this approach, making VLM-based rewards substantially more reliable and practical.
%By addressing these core weaknesses, SAFT moves us one step closer towards the holy grail of fully text-conditioned policies.
Our experiments across both classic control and robotic manipulation environments suggest that VLM reward failures are not always caused by semantic misunderstanding alone, but can also arise from structural brittleness in the reward landscape.
By showing that such brittleness can be mitigated without ground-truth supervision, SAFT supports a broader paradigm in which task structure serves as a scalable source of supervision for practical, fully text-conditioned reinforcement learning.

\bibliographystyle{plainnat}
\bibliography{SAFT}

%%%%%%%%%%%%%%%%%%%%%%%%%%%%%%%%%%%%%%%%%%%%%%%%%%%%%%%%%%%%

\appendix

%Figures and Tables will have S in the name
\renewcommand{\thetable}{S\arabic{table}}
\renewcommand{\thefigure}{S\arabic{figure}}
\renewcommand{\theequation}{S\arabic{equation}}

\section{Implementation details}\label{app:implementation_details}
\subsection{Reinforcement learning setup}
For the MountainCar environment, we use the Stable-Baselines3 \citep{sb3} PPO implementation with default hyperparameters.
For CartPole, Reach, and ReposeCube, we employ the RSL-RL \citep{rudin2022learning} PPO implementation, also with default hyperparameters.
In all environments, the agent receives task reward exclusively from the VLM-based reward model without access to the ground-truth reward.
Exact parameter values can be found in Table~\ref{tab:rl_config}.

\begin{table}[h]
  \caption{Default reinforcement learning parameters from the Stable-Baselines3 \citep{sb3} and RSL-RL \citep{rudin2022learning} PPO implementations, used without modification in our experiments.}
  \label{tab:rl_config}
  \begin{center}
  \begin{small}
      \begin{tabular}{l c c}
        \toprule
        Component & Stable-Baselines3 & RSL-RL \\
        \midrule
        Actor Network & MLP $[64, 64]$ & MLP $[256, 256, 256]$\\
        Critic Network & MLP $[64, 64]$ & MLP $[256, 256, 256]$\\
        Epochs & 10 & 1 \\
        Batch Size & $64$ & \texttt{rollout\_steps}\\
        Learning Rate & $3*10^{-4}$ & $10^{-3}$\\
        Gamma ($\gamma$) & $0.99$ & $0.998$\\
        GAE Lambda ($\lambda$) & $0.95$ & $0.95$\\
        Clip Range & $[-0.2, 0.2]$ & $[-0.2, 0.2]$\\
        Entropy Coefficient & $0$ & $0$\\
        Optimizer & \texttt{Adam} & \texttt{Adam} \\
        \bottomrule
      \end{tabular}
    \end{small}
  \end{center}
  \vskip -0.1in
\end{table}

\subsection{Vision-language model selection and LoRA finetuning}
Following prior work, we adopt the ViT-B-16 backbone from the OpenCLIP \citep{ilharco_gabriel_2021_5143773} collection, pretrained on the LAION-2B dataset \citep{laion2b}, as our base vision-language model.
The vision encoder is fine-tuned once per rollout using LoRA adapters \citep{hu2021loralowrankadaptationlarge}, while the text encoder and all non-LoRA parameters of the vision backbone remain frozen throughout all experiments.
The VLM reward serves as a drop-in replacement for the ground-truth reward, with the textual goal description kept fixed throughout training.
Image observations are encoded by the vision backbone at every step and compared against this fixed goal embedding to produce rewards.
VLM and LoRA configuration details are reported in Table~\ref{tab:lora_config}.

\subsection{Vision-language model pretraining}\label{app:pretraining}

\begin{table}[h]
  \caption{Ground-truth reward definitions for the four environments. CartPole penalizes the pole's angular velocity $\omega_{\text{pole}}$. Reach penalizes the Euclidean distance between the end-effector $\bm{x}$ and the goal $\bm{x}_{\text{goal}}$. MountainCar rewards the car's horizontal position $x$, and punishes large actions $a$. Finally, ReposeCube rewards the agent based on the angular error $\Delta \theta$ between the cube's current and goal quaternions.}
  \label{tab:gt_rewards}
  \begin{center}
  \begin{small}
\begin{tabular}{l c c}
\toprule
Environment & Term & Weight \\
\midrule
\makecell[l]{CartPole} & $-\omega_{\text{pole}}^2$ & $1.0$ \\
\midrule
\makecell[l]{Reach} & $-\| \bm{x} - \bm{x}_{\text{goal}} \|_2$ & $0.2$ \\
& $1 - \tanh\left( \tfrac{\| \bm{x} - \bm{x}_{\text{goal}} \|_2}{\sigma} \right)$ & $0.1$ \\
\midrule
\makecell[l]{MountainCar} & $x + 0.3$ & $1.0$ \\
& $-a^2$ & $0.1$ \\
\midrule
\makecell[l]{ReposeCube} & $\left({\Delta {\theta} + 0.1}\right)^{-1}$ & $1.0$ \\
\bottomrule
\end{tabular}
    \end{small}
  \end{center}
  \vskip -0.1in
\end{table}

To rigorously evaluate SAFT, we must isolate the method's structural contributions from the stochastic quality of the underlying Vision-Language Model.
SAFT is designed to refine and structurally align \emph{existing} reward signals.
Therefore, its utility is naturally a function of the base model's capability.
A base model that outputs pure noise cannot be meaningfully aligned, whereas a model that perfectly reflects the ground truth requires no refinement.

To evaluate SAFT on different points of this spectrum, as presented in Figure~\ref{fig:epic_vs_comparisons}, we employ supervised pretraining as a controlled initialization protocol.
This allows us to smoothly interpolate between these extremes, ranging from random initialization to near-oracle performance, and demonstrate that SAFT provides consistent relative improvements regardless of the base model's starting quality.
This strategy simulates deploying off-the-shelf VLMs across the capability spectrum, treating larger models as stronger initializations, while offering a smooth performance gradient for more precise evaluation than discrete model comparisons.

\textbf{Pretraining methodology} To achieve these targeted initialization states, we pretrain the LoRA adapters by regressing the VLM reward predictions toward the ground-truth task rewards, as listed in Table~\ref{tab:gt_rewards}.
We employ Mean Squared Error (MSE) as the loss function:
\begin{equation}\label{eq:regression}
    L_{\text{pretrain}} = \| r_{\text{VLM}}(o_t) - r_{\text{scaled}}(s_t) \|^2
\end{equation}
where $r_{\text{VLM}}(o_t)$ is the reward output of the VLM, given an observation $o_t$ at time $t$.
Since this reward stems from calculating cosine similarity, and cosine similarity is strictly bounded to $[-1, 1]$, attempting to regress to arbitrary, unconstrained ground-truth values would prevent the VLM from learning the correct reward structure.
To align the targets with the VLM's output space, we linearly scale the ground-truth reward $r_{\text{gt}}(s_t)$, given a state $s$ at time $t$, to the $[-1, 1]$ range:
\begin{equation}
    r_{\text{scaled}}(s_t) = 2 \cdot \frac{r_{\text{gt}}(s_t) - R_{\min}}{R_{\max} - R_{\min}} - 1
\end{equation}
where $R_{\min}$ and $R_{\max}$ represent the empirical minimum and maximum rewards of the environment. During pretraining, we apply this loss after every step.

We select regression over a PbRL-style pretraining objective for this phase because regression allows for linear, low-variance control over the reward structure (measured via EPIC distance).
In contrast, PbRL pretraining tends to yield sparser and noisier rewards \citep{christiano2023deepreinforcementlearninghuman, tao2025hybridreinforcementrewardsparse}, making it difficult to target specific capabilities for controlled evaluation.
Furthermore, we posit that direct regression serves as the most faithful proxy for the behavior of future general-purpose VLMs, as it promotes a representation that remains unbiased toward any specific component of the reward structure.
However, to demonstrate generality, we also provide an evaluation of SAFT applied to a PbRL-pretrained baseline in Appendix~\ref{app:class}.

\textbf{Selection of operating point for main results} While Figure~\ref{fig:epic_vs_comparisons} demonstrates SAFT's utility across a range of model strengths, the detailed learning curves in the main text (Figure~\ref{fig:policy_performance}, Table~\ref{tab:epic_distance}, Table~\ref{tab:comparisons_saved}) utilize a specific initialization threshold where the base VLM is capable of partially solving the task.
We identified this operating point empirically by rolling out the policy from different VLM pretraining checkpoints and visually checking when the policy showed learning and roughly solved the task.
We deliberately selected this functional regime as relying solely on off-the-shelf models that often fail completely, as seen in the ViT-B-16 analysis in Figure~\ref{fig:smaller_model}, would reduce our evaluation to a binary success or failure outcome.
By contrast, evaluating on a functional but imperfect model allows us to demonstrate that SAFT accelerates learning and denoises the reward landscape even when the base model is not catastrophic.

The exact pretraining budget for the base VLM in the main results of Sections~\ref{sec:policy_performance} and \ref{sec:offline_eval} differs by environment. CartPole uses $30$ rollouts, resulting in $30 \times 32 = 960$ individual updates, or in other words $960$ epochs of batch size $1$. Reach uses $48 \times 27 = 1296$. MountainCar uses $100 \times 1$, because a full rollout would be excessive. ReposeCube uses $24 \times 180 = 4320$.

\textbf{Strict separation of phases} We emphasize that this access to ground-truth rewards is strictly limited to the \emph{pretraining phase} to establish experimental starting conditions.
Once the SAFT fine-tuning phase begins, all access to ground truth is revoked, and the model relies exclusively on the self-supervised auxiliary losses ($L_{\text{CAL}}$ and $L_{\text{RLR}}$).
Thus, no ground-truth information leaks into the SAFT update process.

\textbf{Off-the-shelf baselines} Finally, to complement the pretrained experiments, we provide training curves comparing SAFT against an off-the-shelf ViT-B-16 model without any pretraining for all four environments in Figure~\ref{fig:smaller_model}.
These results confirm that even in regimes where the base model fails to converge, SAFT is useful, sometimes even being the deciding factor that recovers a solvable policy.

\begin{table}[h]
\vskip 0.3in
  \caption{Vision-language model and LoRA training parameters used across all experiments.}
  \label{tab:lora_config}
  \begin{center}
  \begin{small}
\begin{tabular}{l c}
\toprule
Component & Configuration \\
\midrule
\multicolumn{2}{l}{\textit{Vision-Language Model}} \\
\quad Backbone & ViT-B-16\\
\quad Parameters & $86$M \\
\quad Source & OpenCLIP, trained on LAION-2B \\
\addlinespace[0.5em]
\multicolumn{2}{l}{\textit{LoRA Adapters}} \\
\quad Parameters & $180$k \\
\quad Target Modules & Feedforward Projection (\texttt{c\_fc}), Output Projection (\texttt{c\_proj})\\
\quad Rank & $2$ \\
\quad Scaling Factor & $32$ \\
\quad Dropout & $0.1$ \\
\quad Bias & None \\
\quad Optimizer & \texttt{Adam} \\
\quad Learning Rate & $10^{-6}$ \\
\bottomrule
\end{tabular}
    \end{small}
  \end{center}
\end{table}

\subsection{Environment-specific configurations and auxiliary loss selection}
\label{app:env_specific_config}

We keep training parameters consistent across environments, introducing only minimal task-specific modifications. Experiments on CartPole, Reach, and ReposeCube are conducted within Isaac Lab \citep{mittal2023orbit}, while MountainCar is run in Gymnasium \citep{brockman2016openaigym}.

The choice of auxiliary loss depends on the inherent structure of each environment. We select between CAL and RLR based on which structural properties can be exploited for reward learning, as summarized in Table~\ref{tab:env_matching}. RLR is used when local state distances provide meaningful information about reward differences, while CAL is used when task-preserving transformations should leave the reward unchanged.

\paragraph{CartPole}
CartPole exhibits both strong state-reward distance correlation and left-right symmetry, since poles tilted \(\pm \theta\) degrees left or right from vertical should receive equal reward. Therefore, CartPole supports both RLR and CAL. In the main SAFT experiments, we use CAL, with horizontal flips as positive samples and vertical flips as hard negatives. For RLR experiments, we use a window size of \(32\).

\paragraph{MountainCar}
MountainCar maintains a state-reward distance correlation suitable for RLR, but lacks exploitable symmetries. Since the ground-truth reward is defined by distance from the right side of the screen, no two distinct states yield equal rewards, and hence no two observations can serve as positive augmentation pairs. We therefore apply only RLR, using a window size of \(500\). We also add an action penalty to stabilize learning and better distinguish the optimal policy from alternatives.

\paragraph{Reach}
In the Isaac Reach task, the camera is mounted above the end effector, showing both current and target markers and centering the view so only relative goal motion is visible. The task is modified to require only position alignment between the end effector and the goal, removing the orientation constraint. Rewards are computed from a top-down projection of proximity rather than full 3D distance, reflecting the limited depth perception of current VLMs. A visual tracker of the target and end effector positions is rendered to the agent.

The circular symmetry around the target, where states at fixed radius have equal reward, violates the assumptions of RLR but provides rotational invariances exploitable by CAL. Reach therefore uses CAL, with \(8\) rotated positive samples at \(\pm 10^{\circ}, \pm 20^{\circ}, \pm 30^{\circ}, \text{and } \pm 40^{\circ}\), along with a soft negative sampled from the same rollout.

\paragraph{ReposeCube}
For ReposeCube, instead of matching a held cube to a goal cube, the task is reformulated to require orienting the cube such that a specific side faces upward, with the cube itself rendered in view. The environment exhibits rotational symmetries similar to Isaac Reach: multiple cube orientations yield equivalent angular distances to the goal, making RLR inappropriate but providing natural invariances for CAL.

ReposeCube therefore uses CAL, with \(6\) rotated positive samples at \(\pm 10^{\circ}, \pm 20^{\circ}, \text{and } \pm 30^{\circ}\), coupled with a hard negative rendered from the opposing \(180^{\circ}\) viewpoint.

Together, these environments cover several structural regimes for VLM reward models, including sensitivity to asymmetric dynamics in CartPole, unique states without valid augmentation pairs in MountainCar, rotational invariances that break distance-based rewards in Reach, and high-dimensional manipulation with non-trivial equivalences across cube orientations in ReposeCube.

After each rollout, we compute one auxiliary loss, either \(L_{RLR}\) or \(L_{CAL}\), depending on the environment, and backpropagate it once. Gradients update only the LoRA adapters in the otherwise frozen vision encoder. To avoid leaking ground-truth information, we disable early stopping in all environments. Depending on task difficulty, we adjust the rollout length and the number of parallel environments. The exact parameters, CAL augmentations, and goal and baseline descriptions used in goal-baseline regularization \citep{rocamonde2024visionlanguage} are provided in Table~\ref{tab:environment_params}.

\begin{table}[h]
\vskip 0.3in
  \caption{Environment-specific parameters and augmentations used in all experiments.
The labels \emph{soft} and \emph{hard} indicate whether the negative corresponds to an imperfect but plausible alternative observation or to a strict contradiction of the current state.}
  \label{tab:environment_params}
  \begin{center}
  \begin{small}
\begin{tabular}{l c c}
\toprule
 & \multicolumn{2}{c}{\textit{Environments}} \\
\cmidrule(lr){2-3}
 & CartPole & MountainCar \\
\addlinespace[0.6em]
Rollout Steps & $32$ & $1\,000$ \\
Parallel Environments & $64$ & $9$ \\
Auxiliary Loss & CAL & RLR \\
CAL Positive Sample & Horizontal Flip & N/A \\
CAL Negative Sample & Vertical Flip (\emph{hard}) & N/A \\
RLR Window Size & 32 & 500 \\
Baseline Prompt & ``pole and cart'' & ``a car in the mountain'' \\
Goal Prompt &
  \makecell[c]{``pole vertically upright \\ on top of the cart''} &
  \makecell[c]{``a car at the peak of the mountain, \\ next to the yellow flag''} \\
%Visualization &
 %\makecell[c]{\vspace{-1.0ex}\\\includegraphics[width=0.2\linewidth]{figures/cartpole_visual.pdf}\\\vspace{-2.0ex}} & 
  %\makecell[c]{\vspace{-1.0ex}\\\includegraphics[width=0.2\linewidth]{figures/mountaincar_visual.pdf}\\\vspace{-2.0ex}} \\
\midrule
 & Reach & ReposeCube \\
\addlinespace[0.6em]
Rollout Steps & $48$ & $24$ \\
Parallel Environments & $64$ & $1\,024$ \\
Auxiliary Loss & CAL & CAL \\
CAL Positive Sample & Rotation ($\pm10/20/30/40^{\circ}$) & Rotation ($\pm10/20/30^{\circ}$) \\
CAL Negative Sample & Past Observation (\emph{soft}) & Opposite Orientation (\emph{hard}) \\
RLR Window Size & N/A & N/A \\
Baseline Prompt & ``a red and a blue object'' & ``a cube with colors'' \\
Goal Prompt &
  \makecell[c]{``the red and blue object are \\ in the same place''} &
  \makecell[c]{``only the pink side of the \\ cube is visible''} \\
%Visualization &
%\makecell[c]{\vspace{-1.0ex}\\\includegraphics[width=0.2\linewidth]{figures/reach_visual.pdf}\\\vspace{-2.0ex}} & \makecell[c]{\vspace{-1.0ex}\\\includegraphics[width=0.2\linewidth]{figures/reposecube_visual.pdf}\\\vspace{-2.0ex}} \\
\bottomrule
\end{tabular}
    \end{small}
  \end{center}
  \vskip -0.1in
\end{table}

\subsection{PbRL baseline}\label{app:pbrl_details}

To benchmark human labeling efficiency, as done in Section~\ref{sec:human_labeling_efficiency}, we implement a standard Preference-based Reinforcement Learning (PbRL) baseline.
Instead of real-time human feedback, we employ a synthetic oracle derived from the ground-truth reward functions defined in Table~\ref{tab:gt_rewards}.
We generate query pairs by sequentially pairing observations $(o_t, o_{t+1})$ from the rollout buffer as they are collected.
The oracle provides a binary preference $y \in \{0, 1\}$ indicating which observation corresponds to a higher ground-truth reward.
The reward model is updated online by minimizing the negative log-likelihood under the Bradley-Terry model (Equation~\ref{eq:pbrl_1}).
Consistent with the SAFT setup, we freeze the VLM backbone and only update the LoRA adapters described in Table~\ref{tab:lora_config}.

\subsection{EPIC distance calculation}
EPIC distance, as defined by \cite{gleave2021quantifyingdifferencesrewardfunctions}, is evaluated over a distribution of observations and their true rewards.
To prevent skew within this distribution and improve the validity of the evaluation, we collect half of the observations using a random policy, and the other half from a policy trained using the true reward.
In total, $2\,000$ observations are collected for evaluation.
For ReposeCube, we directly render the cube, enabling us to synthetically generate $10\,000$ random rotations for EPIC distance evaluation.

\subsection{Training infrastructure and runtime}\label{app:hardware}

All experiments were conducted using two NVIDIA A100 GPUs with 80GB memory each. 
The additional fine-tuning performed by SAFT introduces a wall-clock overhead compared to the base VLM reward model, as LoRA updates are performed online during policy training.

We report the corresponding training times for the main experiments in Section~\ref{sec:policy_performance} in Table~\ref{tab:training_time}. These results provide a direct comparison between the base reward model and SAFT across all environments.

\begin{table}[h]
  \caption{Wall-clock training time for the base VLM reward model and SAFT across environments from Section~\ref{sec:policy_performance}.}
  \label{tab:training_time}
  \begin{center}
  \begin{small}
      \begin{tabular}{l c c c c}
        \toprule
        Method & CartPole & MountainCar & Reach & ReposeCube \\
        \midrule
        Base & 2\,h & 1\,h & 10\,h & 29\,h \\
        SAFT & 4\,h & 2\,h & 22\,h & 63\,h \\
        \bottomrule
      \end{tabular}
    \end{small}
  \end{center}
  \vskip -0.1in
\end{table}

\section{Additional results}

\subsection{SAFT enables off-the-shelf use of smaller models}\label{app:smaller_models}

As discussed in Section~\ref{sec:human_labeling_efficiency}, there exists a regime in which the underlying VLM alone cannot solve the task but succeeds when using SAFT.

\begin{figure}[t]
    \begin{center}
    \centerline{\includegraphics[width=1.0\textwidth]{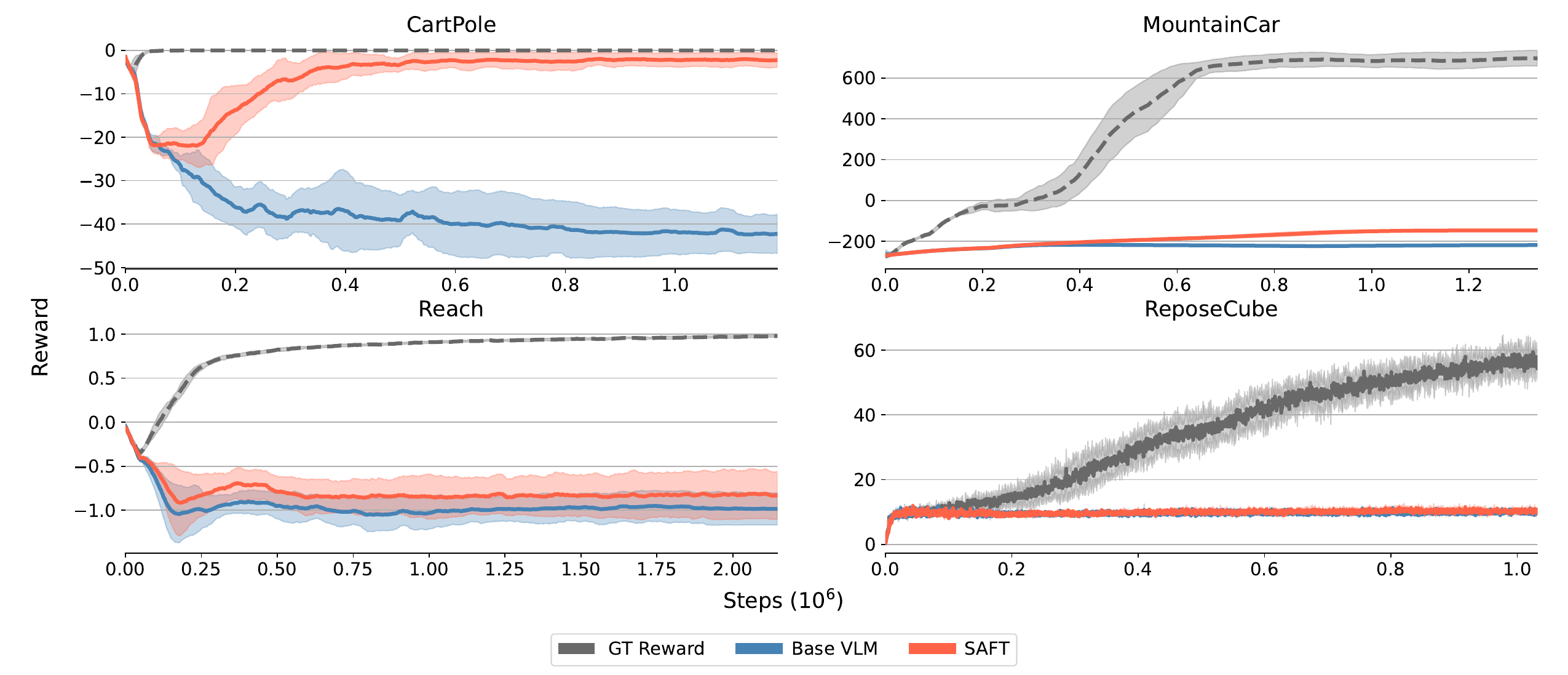}}
    \caption{Results are averaged over five seeds for all environments, except ReposeCube which uses three.
    The off-the-shelf ViT-B-16 model fails to converge without SAFT, which shows that SAFT can be essential for convergence in smaller models.}
    \captionsetup{justification=centering}
    \label{fig:smaller_model}
    \end{center}
    \vskip -0.2in
\end{figure}

Figure~\ref{fig:smaller_model} illustrates the effects of SAFT when applied to an off-the-shelf ViT-B-16 model \citep{ilharco_gabriel_2021_5143773}.
We can observe that for the CartPole environment, the underlying model fails to converge unless SAFT is applied.
This result highlights that enforcing the relative structure within the reward model via SAFT is not only beneficial for improving sample efficiency in stronger models (Figure~\ref{fig:policy_performance}), but can also determine whether the training converges at all when using weaker ones. 

\subsection{Necessity of fine-tuning}

\begin{figure}
    \begin{center}
    \centerline{\includegraphics[width=1.0\textwidth]{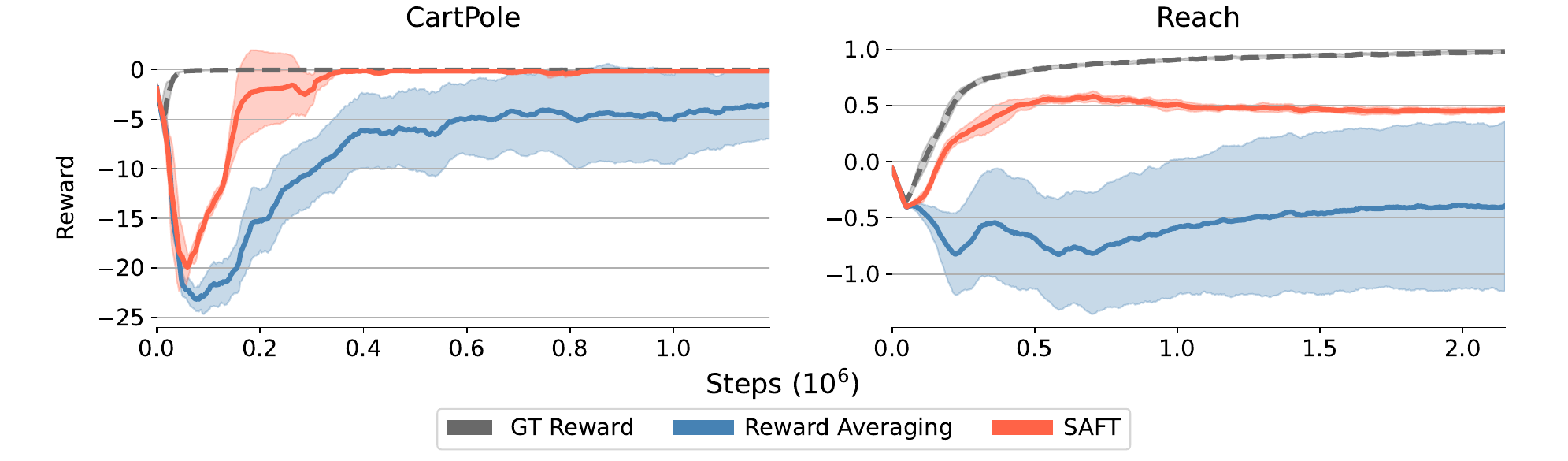}}
    \caption{Comparison of SAFT with naive reward averaging over positive augmentations. Fine-tuning consistently outperforms averaging, indicating that simple smoothing falls short. Results presented over five seeds.}
    \captionsetup{justification=centering}
    \label{fig:necessity_of_finetuning}
    \end{center}
    \vskip -0.2in
\end{figure}

As fine-tuning the VLM incurs computational overhead, we first ask whether it offers benefits beyond a simple smoothing strategy that does not require any backpropagation.
We compare SAFT, using CAL, to a naive baseline that averages the VLM’s predicted rewards across positive augmentations without any fine-tuning.
Figure~\ref{fig:necessity_of_finetuning} shows that fine-tuning consistently outperforms reward averaging, indicating that negative augmentations provide essential signal and that fine-tuning lets the model correct rewards for nearby states within the state space.
This suggests that the gains come from learning rather than mere smoothing, and that the additional compute is justified.

\subsection{Weighting of CAL terms}\label{app:cal_weighting}

\begin{figure}[t]
    \begin{center}
    \centerline{\includegraphics[width=1.0\textwidth]{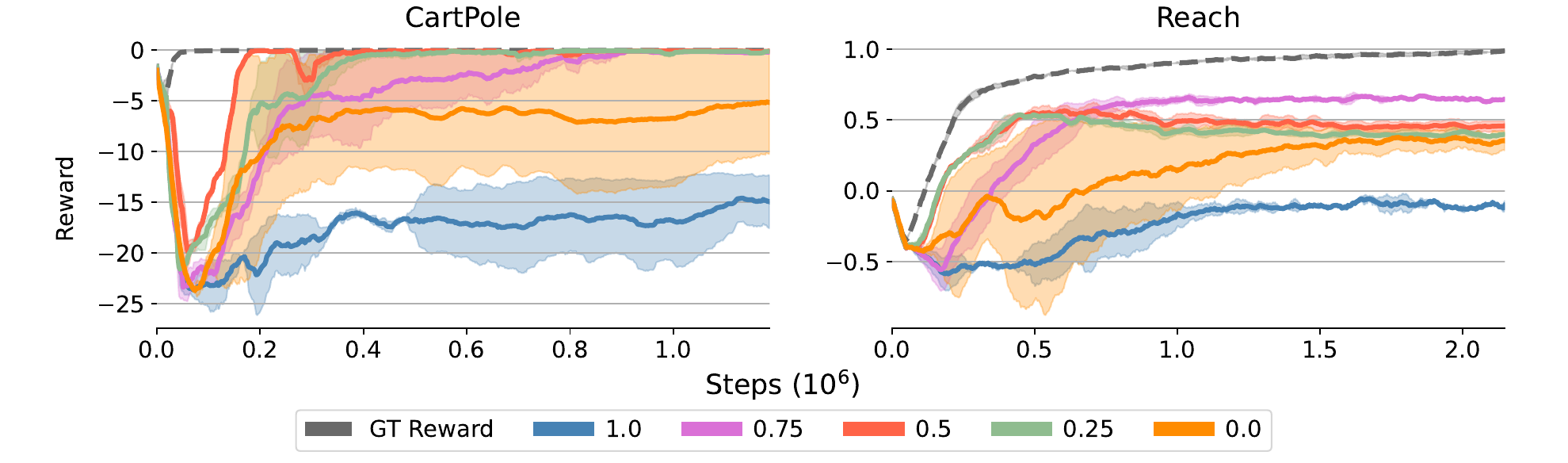}}
    \caption{Ablation of the CAL weighting term $\beta$ between the positive-consistency and negative-separation terms, averaged over two seeds.
    Results indicate complementary contributions, with performance degrading when either component is downweighted.}
    \captionsetup{justification=centering}
    \label{fig:cal_weighting}
    \end{center}
    \vskip -0.2in
\end{figure}

The CAL objective has two components, a positive consistency term that minimizes variation in rewards across positive transformations, and a negative separation term that maximizes variation against negative examples.
To assess the contribution of each term, we ablate the weighting parameter $\beta$ in Equation~\ref{eq:CAL} by comparing our default $\beta = 0.5$ against $\beta \in \{1.0, 0.75, 0.25, 0.0\}$.
The results presented in Figure~\ref{fig:cal_weighting} show that both terms contribute, with the negative separation term preventing representational collapse and the positive consistency term providing additional gains.

\subsection{RLR window size}\label{app:rlr_window_size}

\begin{figure}[t]
    \begin{center}
    \includegraphics[width=1.0\textwidth]{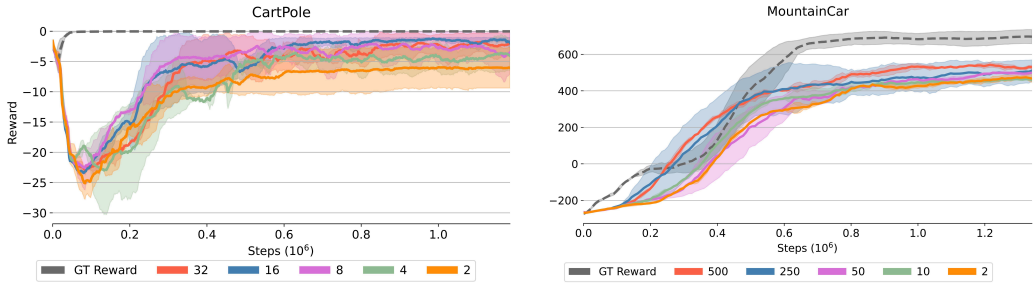}
    \caption{Ablation of the RLR window size $W$ determining the strength of the temporal consistency prior, averaged over two seeds.
    Results indicate that for both environments, larger window sizes tend to fare better.}
    \captionsetup{justification=centering}
    \label{fig:rlr_window_size}
    \end{center}
    \vskip -0.2in
\end{figure}

The window size $W$ modulates the RLR temporal consistency prior, directly scaling the loss strength.
We evaluate the sensitivity to this parameter by comparing our default settings against reduced window sizes: $W \in \{16, 8, 4, 2\}$ for CartPole (default $W=32$) and $W \in \{250, 50, 10, 2\}$ for MountainCar (default $W=500$).

Figure~\ref{fig:rlr_window_size} demonstrates that larger window sizes improve performance in both environments. This suggests that the temporal consistency prior, which relates observation-space distances to reward-space distances, remains robust over long horizons.

\subsection{Simultaneous usage of CAL and RLR}

\begin{figure}[t]
    \begin{center}
    \includegraphics[width=0.8\textwidth]{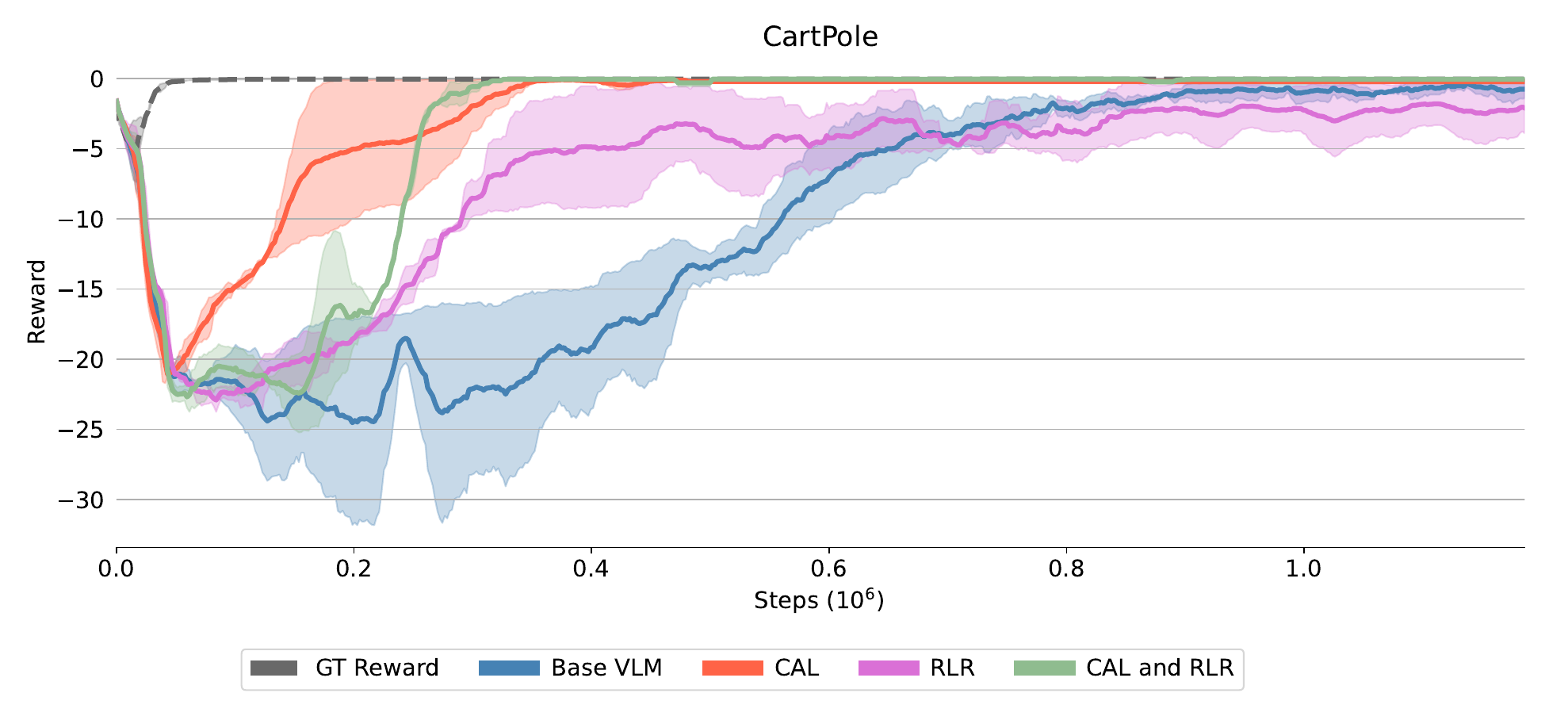}
    \caption{CartPole results comparing CAL, RLR, and their combination, averaged over two seeds. CAL outperforms RLR, and while each method is effective individually, their combination provides no additional gain.}
    \captionsetup{justification=centering}
    \label{fig:cal_vs_rlr}
    \end{center}
    \vskip -0.2in
\end{figure}

The CartPole environment is unique in that it supports both RLR and CAL (Table~\ref{tab:env_matching}).
This lets us compare them directly and test whether their effects stack.
Figure~\ref{fig:cal_vs_rlr} shows that CAL outperforms RLR, likely because its positive and negative transformations are both precise and unique.
Notably, although both methods are helpful on their own, their combination yields no further improvement.

%\hl{Appendix ablations TODO:}
%\hl{- maybe add ViT-bigG-14 runs (cartpole)}
%\hl{- reach more rotations results (show it isnt hand-tuned) (reach)}

\subsection{Pretraining using preference-based reinforcement learning}\label{app:class}

\begin{figure}[t]
    \begin{center}
    \includegraphics[width=0.95\textwidth]{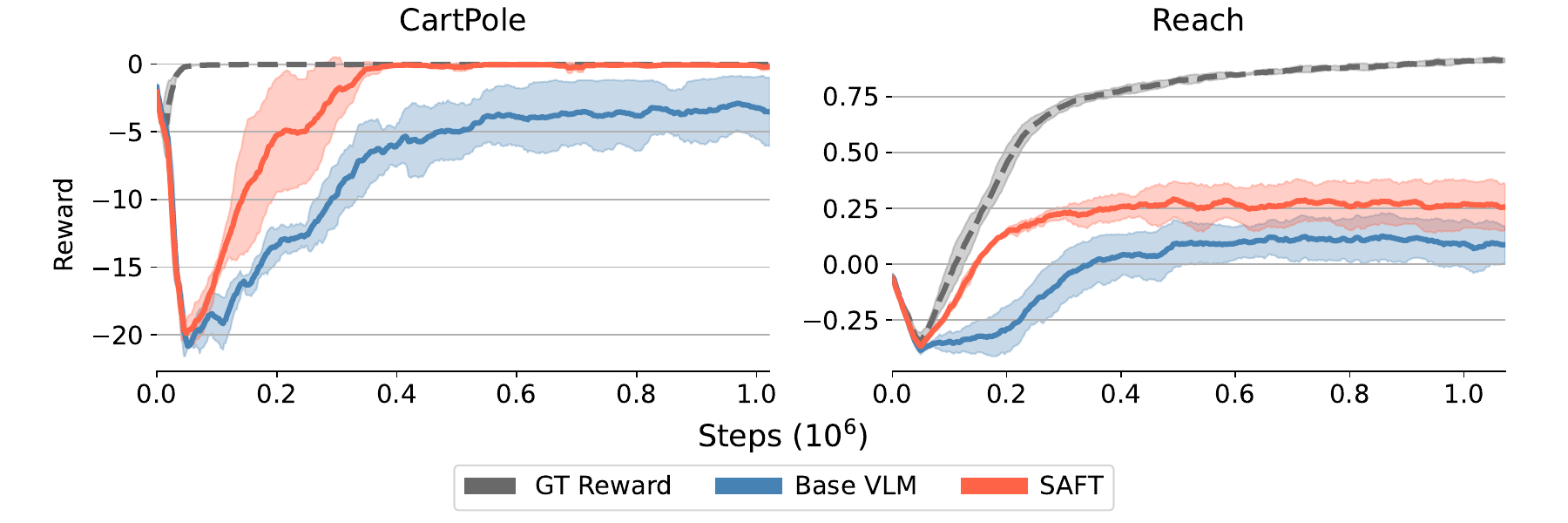}
    \caption{Application of SAFT to a model pretrained using a PbRL objective, averaged over five seeds.
    Results demonstrate that SAFT also yields significant performance gains following PbRL pretraining.}
    \captionsetup{justification=centering}
    \label{fig:class}
    \end{center}
    \vskip -0.2in
\end{figure}

As an alternative to the regression loss in Equation~\ref{eq:regression}, we employ PbRL pretraining for cases where the initial VLM quality precludes solving the task. This approach requires no ground-truth rewards yet effectively primes the model for SAFT, as shown in Figure~\ref{fig:class}. We utilize the standard Bradley-Terry formulation of PbRL from Equations~\ref{eq:pbrl_1} and \ref{eq:pbrl_2}, performing updates after each rollout.

We note that the observed gains are smaller than those in Sections~\ref{sec:policy_performance} and \ref{sec:offline_eval}.
This is consistent with Section~\ref{sec:human_labeling_efficiency}, as the pretraining phase was longer and yielded a more capable base model.

\subsection{Additional backbone experiments}\label{app:backbone}

\begin{figure}[t]
    \begin{center}
    \hspace*{0.02\textwidth}
    \includegraphics[width=0.8\textwidth]{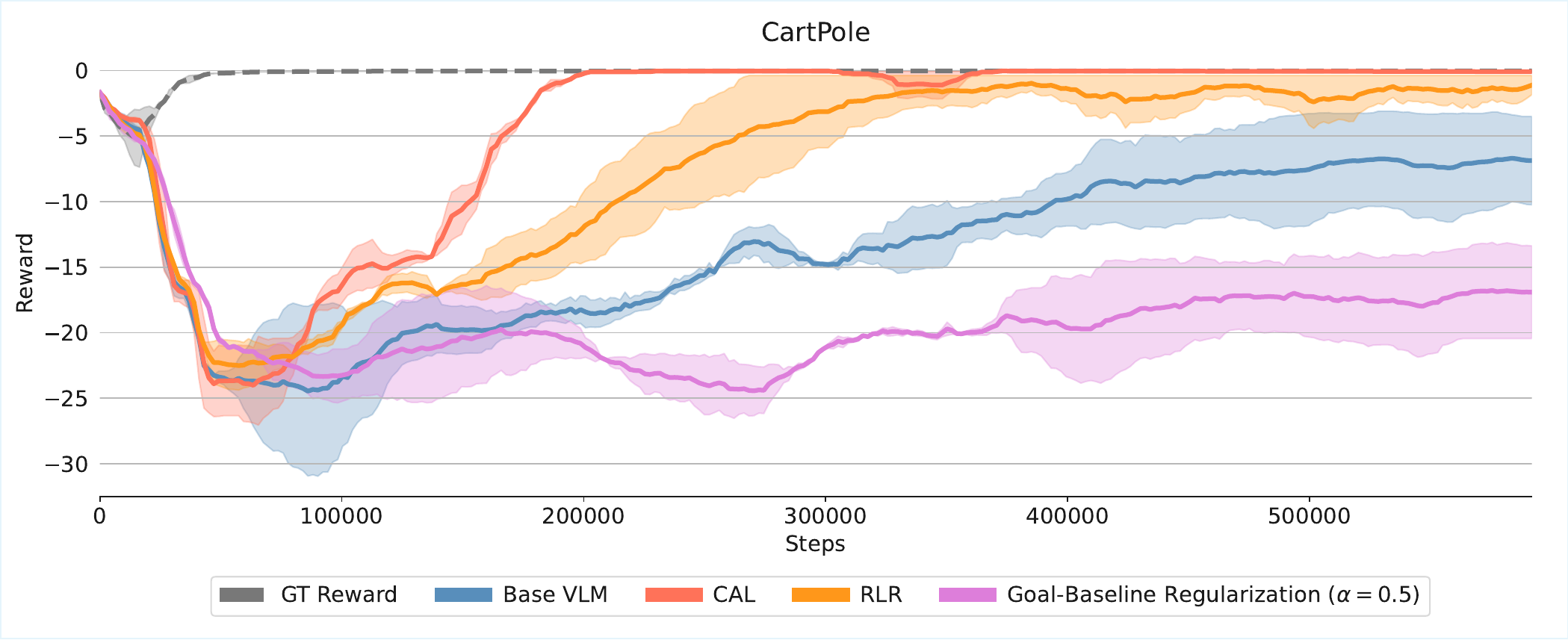}
    \caption{Application of SAFT to a similarly-sized SigLIP2 \citep{tschannen2025siglip2multilingualvisionlanguage} backbone, pretrained under the same conditions as Section~\ref{sec:policy_performance}, averaged over two seeds.
    Results demonstrate that SAFT consistently yields significant performance gains across backbones.}
    \captionsetup{justification=centering}
    \label{fig:siglip}
    \end{center}
    \vskip -0.2in
\end{figure}

\begin{figure}[t]
    \begin{center}
    \hspace*{0.02\textwidth}
    \includegraphics[width=0.8\textwidth]{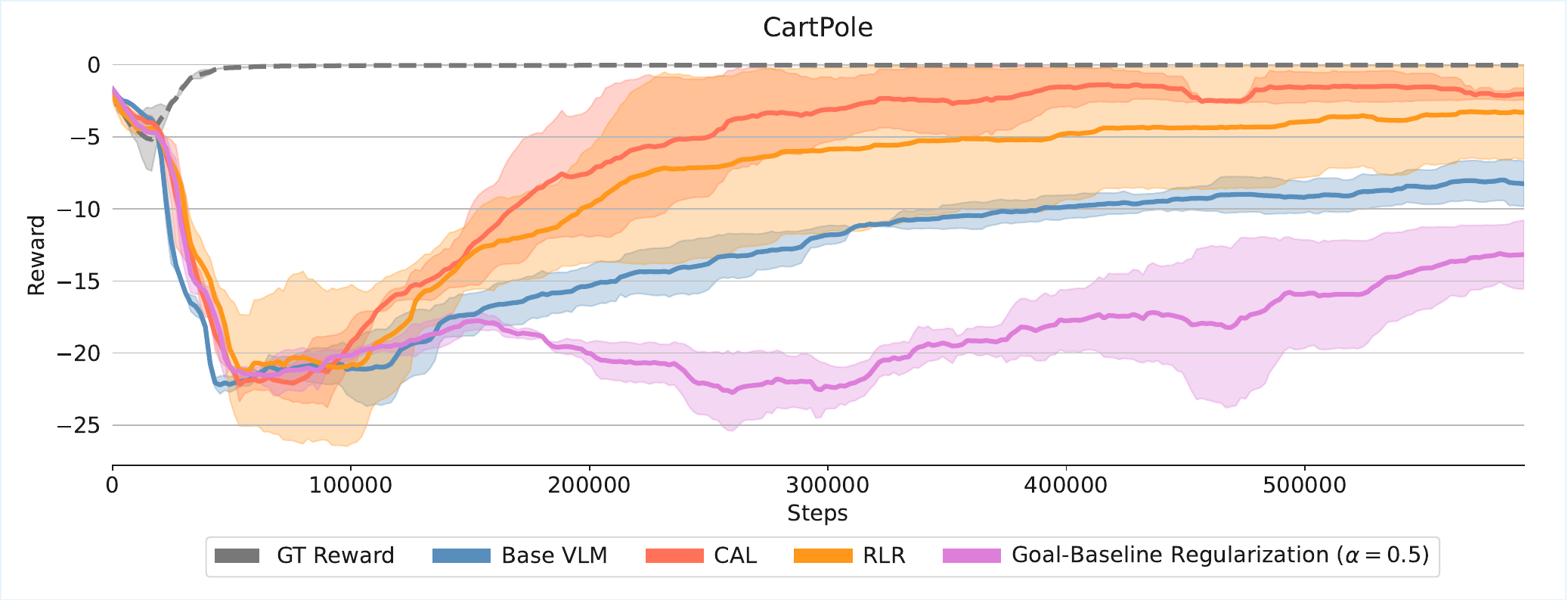}
    \caption{Application of SAFT to a similarly-sized MetaCLIP \citep{chuang2025metaclip2} backbone, pretrained under the same conditions as Section~\ref{sec:policy_performance}, averaged over two seeds.
    Results demonstrate that SAFT consistently yields significant performance gains across backbones.}
    \captionsetup{justification=centering}
    \label{fig:metaclip}
    \end{center}
    \vskip -0.2in
\end{figure}

Although there is no clear theoretical reason for SAFT to behave differently across architectures, as it operates via LoRA updates on the image encoder and focuses on improving reward structure rather than model-specific components, we additionally evaluate it using similarly-sized SigLIP2 \citep{tschannen2025siglip2multilingualvisionlanguage} and MetaCLIP \citep{chuang2025metaclip2} backbones (\texttt{siglip2-base-patch16-224}, \texttt{
metaclip-2-worldwide-s16-384}).

Training is performed under the exact same setup as the main experiments in Section~\ref{sec:policy_performance}, including identical environment configurations, reward formulation, and LoRA parameterization (fine-tuning approximately $0.2\%$ of total parameters), with no additional tuning or modifications.

We report results on CartPole, as this environment allows direct comparison of both CAL and RLR within a single setting. Across two seeds, Figure~\ref{fig:siglip} and Figure~\ref{fig:metaclip} show that SAFT consistently improves performance over both the base model and goal-baseline regularization.
Both CAL and RLR variants yield clear gains, confirming that the improvements from SAFT are not specific to a particular backbone.

%%%%%%%%%%%%%%%%%%%%%%%%%%%%%%%%%%%%%%%%%%%%%%%%%%%%%%%%%%%%

%\newpage
%\input{checklist.tex}

\end{document}